\pdfoutput=1
\documentclass[11pt]{article}

\usepackage{acl}

\usepackage{times}
\usepackage{latexsym}

\usepackage[T1]{fontenc}
\usepackage[utf8]{inputenc}

\usepackage{microtype}
\usepackage{graphicx}
\usepackage{multirow}
\usepackage{amsmath}
\usepackage{booktabs}
\usepackage{microtype}
\usepackage{hyperref}
\usepackage{amsfonts}
\usepackage{bbding}
\usepackage{color}
\title{GroundingGPT: Language Enhanced Multi-modal Grounding Model}

\author{
    \textbf{Zhaowei Li} \\
    ByteDance Inc\\
}
\author{
    \textbf{Zhaowei Li\textsuperscript{\rm 1,2}},
    ~\textbf{Qi Xu\textsuperscript{\rm 1}},
    ~\textbf{Dong Zhang\textsuperscript{\rm 2}},
    ~\textbf{Hang Song\textsuperscript{\rm 1}},
    ~\textbf{Yiqing Cai\textsuperscript{\rm 1}},\\
    ~\textbf{Qi Qi\textsuperscript{\rm 1}},
    ~\textbf{Ran Zhou\textsuperscript{\rm 1}},
    ~\textbf{Junting Pan\textsuperscript{\rm 1}},
    ~\textbf{Zefeng Li\textsuperscript{\rm 1}},
    ~\textbf{Van Tu Vu\textsuperscript{\rm 1}},\\
    ~\textbf{Zhida Huang\textsuperscript{\rm 1}},
    ~\textbf{Tao Wang\textsuperscript{\rm 1}}\\
    \textsuperscript{\rm 1}ByteDance Inc, \textsuperscript{\rm 2}Fudan University\\
    {\tt lizhaowei126@gmail.com} \\
    \url{https://lzw-lzw.github.io/GroundingGPT.github.io/}
}
\begin{document}
\twocolumn[{
\renewcommand\twocolumn[1][]{#1}
\maketitle
% \begin{center}
%     \captionsetup{type=figure}
%     \includegraphics[width=\textwidth]{Figures/alldemo.pdf}
%     \captionof{figure}{GroundingGPT is an end-to-end unified multi-modal grounding model. We showcase the performance of GroundingGPT across a range of multi-modal tasks, including: 1)Image Grounding, 2)Video Grounding, 3)Sound Localization, 4)Multi-modal Understanding.}
% \end{center}
}]

% \begin{figure*}[h]
%     \centering
%     \includegraphics[width=\textwidth]{Figures/alldemo.pdf}
%     \caption{GroundingGPT is an end-to-end unified multimodal grounding model. We demonstrate GroundingGPT's performance on various multimodal tasks:1)Imgae Grounding;2)Video Grounding;3)Sound Localization;4)Multimodal Understanding.}
%     \label{fig:capability}
% \end{figure*}
\begin{abstract}
Multi-modal large language models (MLLMs) have demonstrated remarkable performance across various tasks. However, these models often prioritize capturing global information and overlook the importance of perceiving local information. This limitation hinders their ability to effectively understand fine-grained details and handle grounding tasks that necessitate nuanced comprehension. 
Although some recent works have made strides in this, they have primarily focused on single-modality inputs.
Therefore, we propose \textbf{GroundingGPT}, an end-to-end language enhanced multi-modal grounding model. It is designed to perform fine-grained grounding tasks for three modalities: image, video and audio. 
% To enhance the model's performance, we introduce 1) coarse-to-fine training strategy: adopt the three-stage training to gradually enhance the model's semantic awareness and fine-grained understanding capabilities;  2) diversified dataset construction pipeline: utilize stage-specific pipeline to develop a multi-modal, multi-granularity dataset for model training in different stages.
To enhance the model's performance, we adopt a coarse-to-fine training strategy, utilizing a three-stage training approach to progressively enhance the model's semantic awareness and fine-grained understanding capabilities. Additionally, we employ a diversified stage-specific dataset construction pipeline, developing a multi-modal, multi-granularity dataset tailored for training the model in different stages.
Extensive experiments conducted on multiple multi-modal benchmarks demonstrate that our model achieves impressive fine-grained understanding of multi-modal inputs on grounding tasks while maintaining or improving its global comprehension capabilities. We will make the code, dataset, and model publicly available to facilitate further research in this area.

\end{abstract}

\section{Introduction}
% Significant advancements have recently been made in the field of large language models (LLMs), which have demonstrated superior performance in a variety of natural language processing tasks~\cite{touvron2023llama,zeng2022glm}. 
Building upon the capabilities of large language models (LLMs), research on multi-modal large language models (MLLMs) has also advanced, enabling understanding across a broader range of modalities. Representative models such as LLaVA~\cite{liu2023visual} and MiniGPT-4~\cite{zhu2023minigpt} align visual features obtained from image encoders with LLM embedding space through visual instruction tuning, facilitating tasks such as image captioning and visual question answering.

However, existing MLLMs primarily focus on capturing global information while neglecting the fine-grained local information in multi-modal inputs. This limitation restricts their applicability in grounding tasks requiring a more detailed understanding. 
Shikra~\cite{chen2023shikra}, BuboGPT~\cite{zhao2023bubogpt} and Ferret~\cite{you2023ferret} have explored techniques that enable finer alignment and understanding of inputs.
By considering local-level information, these models exhibit enhanced performance in grounding or referring tasks. These methods provide insights into fine-grained understanding, but they are primarily limited to a single modality. 
There is still significant potential for exploring fine-grained understanding across other modalities.
\begin{table}[t]
    \centering
    \setlength{\tabcolsep}{1mm}{
    \begin{tabular}{c ccc cc}
    \toprule
    \multirow{2}{*}{Models} & \multicolumn{3}{c}{Grounding Modality} & \multirow{2}{*}{Multi} & \multirow{2}{*}{E2E}\\
    \cline{2-4}
    &  Image & Video & Audio &  & \\
    \midrule
    LLaVA           & \textcolor{red}{\XSolidBrush} & \textcolor{red}{\XSolidBrush} & \textcolor{red}{\XSolidBrush} & \textcolor{green}{\CheckmarkBold} & \textcolor{green}{\CheckmarkBold}  \\
    Video-LLaMA     & \textcolor{red}{\XSolidBrush} & \textcolor{red}{\XSolidBrush} & \textcolor{red}{\XSolidBrush} & \textcolor{green}{\CheckmarkBold} & \textcolor{green}{\CheckmarkBold}  \\
    Shikra          & \textcolor{green}{\CheckmarkBold} & \textcolor{red}{\XSolidBrush} & \textcolor{red}{\XSolidBrush} & \textcolor{red}{\XSolidBrush} & \textcolor{green}{\CheckmarkBold}   \\
    Ferret          & \textcolor{green}{\CheckmarkBold} & \textcolor{red}{\XSolidBrush} & \textcolor{red}{\XSolidBrush}  & \textcolor{green}{\CheckmarkBold} & \textcolor{red}{\XSolidBrush}  \\
    BuboGPT         & \textcolor{green}{\CheckmarkBold} & \textcolor{red}{\XSolidBrush} & \textcolor{green}{\CheckmarkBold} & \textcolor{green}{\CheckmarkBold} &  \textcolor{red}{\XSolidBrush}   \\     
    \midrule
    GroundingGPT    & \textcolor{green}{\CheckmarkBold} &  \textcolor{green}{\CheckmarkBold} &  \textcolor{green}{\CheckmarkBold} &  \textcolor{green}{\CheckmarkBold} &  \textcolor{green}{\CheckmarkBold}   \\
    \bottomrule
    \end{tabular}}
    \caption{
    Comparison of multi-modal large language models. 
    "Grounding Modality" refers to the modalities in which the model is capable of performing grounding tasks.
    "Multi" refers to the model's ability to engage in multi-turn conversations with users.
    "E2E" refers to the models that are designed to be end-to-end architecture without the need for external modules.}
    \label{tab:comparison}
\end{table}

To address the aforementioned issue, this paper proposes \textbf{GroundingGPT}, a language enhanced multi-modal grounding model, which is an end-to-end unified large language model designed to perform multi-modal grounding and understanding tasks across various modalities, including image, video, and audio. The comparison between our model and other models can be found in Table \ref{tab:comparison}. Specifically, our model employs modality-specific adapters to map feature representations from individual encoders to the embedding space of LLMs. To incorporate spatial and temporal information, we directly represent coordinates and timestamps as textual numbers, eliminating the need for vocabulary expansion. 
For training GroundingGPT, we design a three-stage coarse-to-fine training strategy. In the first stage, we align each pre-trained multi-modal encoder with the LLM embedding space using modality-specific adapters. In the second stage, we aim to enable the model to capture fine-grained information, including coordinates and timestamps. In the third stage, we perform multi-granularity instruction tuning to refine the model's responses. 
For each stage, we employed a stage-specific dataset construction pipeline to generate a diverse, multi-modal, and multi-granularity training dataset. 
% Acquiring fine-grained multi-modal instruction data poses significant challenges. To overcome this obstacle, we devise a dataset construction pipeline to construct our training dataset. This pipeline incorporates diverse construction methods that are specifically tailored to different data sources, ensuring comprehensive coverage of various scenarios that involve multi-modal interactions. 

To summarize, our contributions are as follows:
\begin{itemize}
    \item We propose GroundingGPT, an end-to-end multi-modal grounding model that accurately comprehends inputs and possesses robust grounding capabilities across multi modalities, including image, video and audio. To the best of our knowledge, GroundingGPT is the first model to achieve multi-modal fine-grained understanding and grounding.
    \item For training GroundingGPT, we employ a three-stage coarse-to-fine training process that enables the model to capture high-level semantic information and low-level fine-grained details simultaneously.
    To address the issue of limited data, we construct a diverse and high-quality multi-modal training dataset, which comprises a rich collection of multi-modal data enriched with fine-grained information. 
    \item Extensive experiments conducted on a wide range of MLLM benchmarks demonstrate the generality and effectiveness of GroundingGPT in multi-modal grounding and understanding tasks across various modalities. 
\end{itemize}

\begin{figure*}[t]
    \centering
    \includegraphics[width=0.95\textwidth]{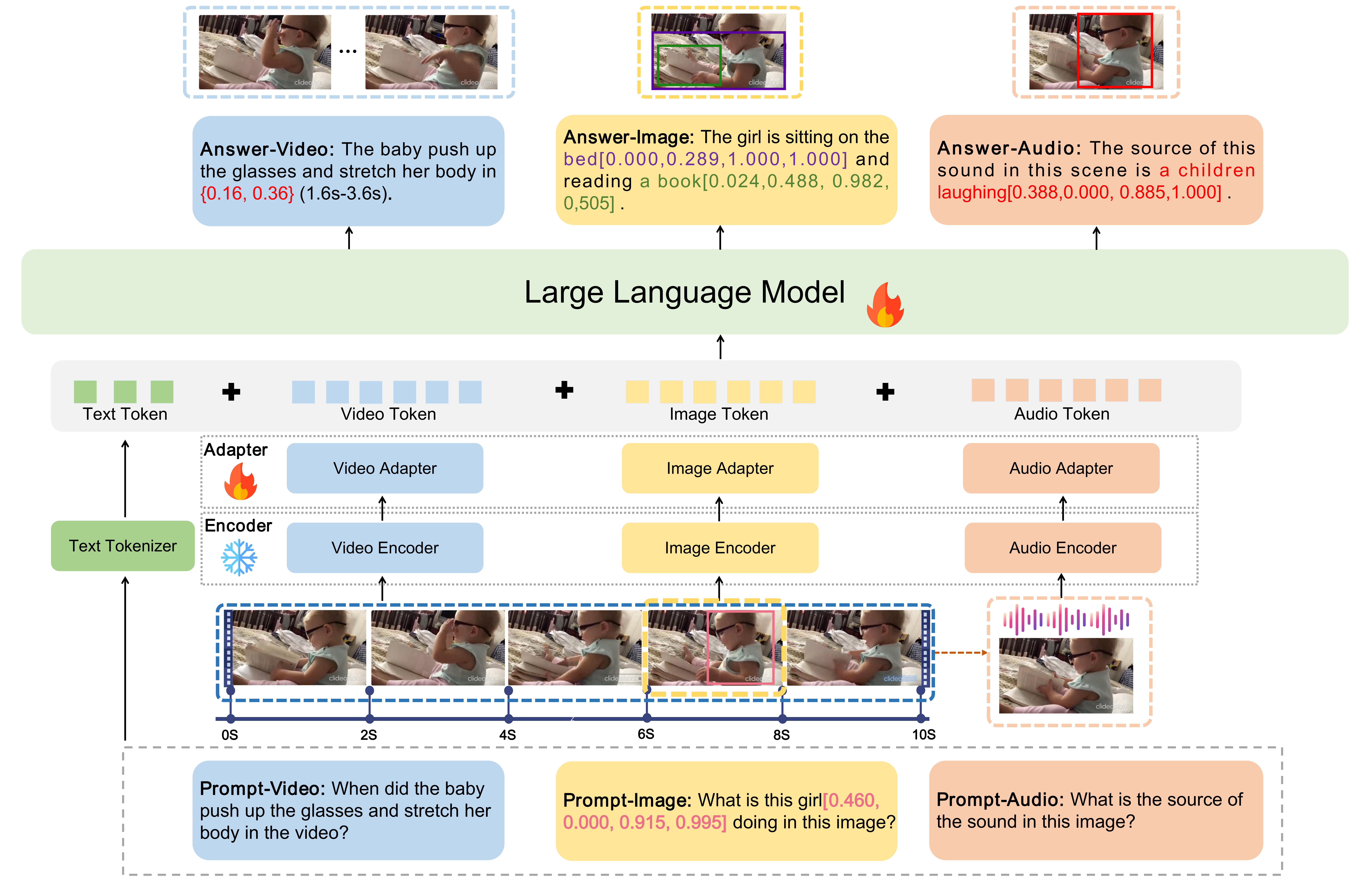}
    \caption{The overall structure of GroundingGPT involves separate encoders and adapters for each modality. Blue boxes represent video inputs, yellow boxes represent image inputs, and pink boxes represent audio inputs.}
    \label{fig:architecture}
\end{figure*}
\section{Related Work}
\label{sec:headings}
\paragraph{Multi-modal Large Language Models (MLLMs)}
Recently, large language models (LLMs) represented by GPTs~\cite{brown2020language,OpenAI2023} and LLaMA~\cite{touvron2023llama} have received extensive attention from researchers for their remarkable performance in various natural language processing tasks.
Substantial progress has been made in the field of MLLMs, which extend the support for multi-modal input and output beyond language. 
These MLLMs typically fine-tune pre-trained LLMs with multi-modal instructions, to enable understanding across multiple modalities. Models such as LLaVA, MiniGPT-4, and mPLUG-Owl~\cite{ye2023mplugowl} map image embeddings obtained from image encoders into the LLM space. Similarly, video MLLMs like Video-Chat~\cite{2023videochat}, Video-LLaMA~\cite{zhang2023video}, Video-Chatgpt~\cite{maaz2023video} and Valley~\cite{luo2023valley}, as well as speech MLLMs like SpeechGPT\cite{zhang2023speechgpt} and LLaSM~\cite{shu2023llasm}, acquire multi-modal understanding capabilities through similar approaches. 
In X-LLM~\cite{chen2023x}, each modality is processed independently through dedicated branches for multi-modal input processing. Pandagpt~\cite{su2023pandagpt} employs a unified embedding space trained by ImageBind~\cite{girdhar2023imagebind} to facilitate joint understanding of various modal inputs. 
% On the other hand, Next-GPT~\cite{wu2023next} achieves both multi-modal input and output by connecting different modality-specific diffusion models at the output end. 
However, these models often fail to adequately capture details within inputs.
% Consequently, their performance may be suboptimal when tackling tasks that require a more detailed understanding.

\paragraph{MLLMs For Grounding Task}
Recently, there has been a focus on training visual MLLMs to achieve fine-grained image understanding and visual grounding. Approaches such as KOSMOS-2~\cite{peng2023kosmos} and Shikra achieve this by incorporating coordinates into the training data, enabling MLLMs to understand the location within images.
On the other hand, approaches like NExT-Chat~\cite{zhang2023next}, GlaMM~\cite{rasheed2023glamm} and Ferret enhance perception of fine-grained information by introducing additional region encoder modules. VTimeLLM~\cite{huang2023vtimellm} demonstrates the capability to understand fine-grained video moment and reason with respect to time boundary.
% These advancements demonstrate the effort made to incorporate fine-grained information into MLLMs, enabling them to achieve more detailed understanding and grounding across different modalities.
% However, the aforementioned models are limited to a single modality, there is still a need for further exploration of fine-grained understanding in other modalities such as video and audio. 
BuboGPT~\cite{zhao2023bubogpt} enables cross-modal interaction between image, audio, and language, facilitating fine-grained understanding of different modalities. 
% For video inputs, PG-Video-LLaVA\cite{munasinghe2023pg} is a MLLM with video localization capabilities. It incorporates existing grounding models and a modular design to successfully accomplish video grounding tasks.
% In contrast, our proposed model supports the understanding of multi-modal information at different granularities, with a unified end-to-end structure. 
% It can be applied to complex multi-modal interactive tasks such as image grounding, video temporal grounding. To the best of our knowledge, this is the first large language model that achieves multi-modal and fine-grained perception across modalities.

\section{Methods}
% This section provides an overview of the model structure, highlighting the branches for different modalities. We also introduce our spatial-temporal representation method and the pipeline used to construct the multi-modal dataset. Additionally, we describe the three-stage coarse-to-fine training process employed by the GroundingGPT model.
We introduce the overall architecture of the GroundingGPT model in this section. Additionally, we will present our three-stage coarse-to-fine training strategy and data construction pipeline.
\begin{figure*}[t]
    \centering
    \includegraphics[width=0.9\textwidth]{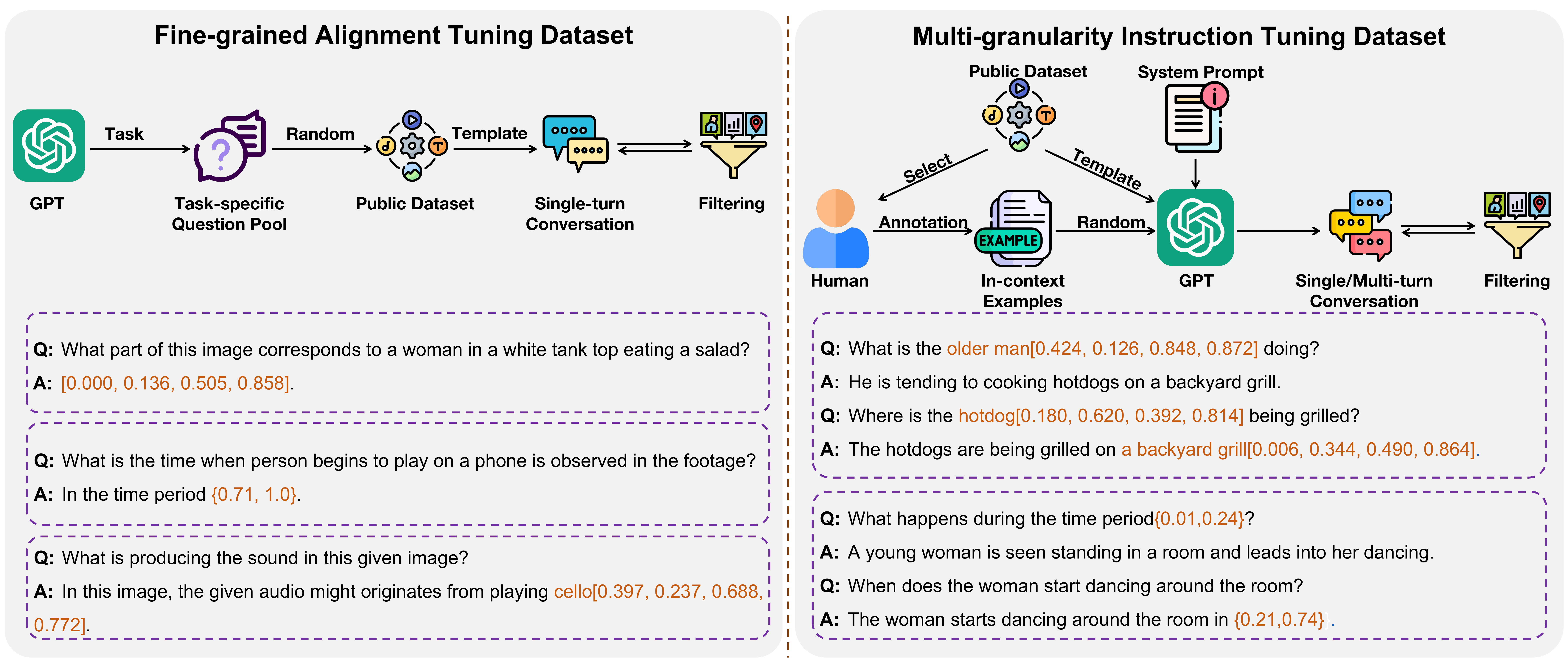}
    \caption{The data construction pipeline and examples for the last two training stages. To simplify, the multi-turn conversation examples only showcase two rounds of question-answer interactions.}
    \label{fig:dataset}
\end{figure*}

\subsection{Model Architecture}
Figure~\ref{fig:architecture} illustrates the overall architecture of the GroundingGPT model. Multi-modal inputs are processed through modality-specific encoders to extract features. These features are then mapped to the LLM embedding space using corresponding adapters. We will also introduce the representation of coordinates and timestamps.

\subsubsection{Image Branch}
We employ the pre-trained CLIP visual encoder ViT-L/14~\cite{radford2021learning} to extract image features. 
% Similar to ~\cite{liu2023visual}, we select the features before the last Transformer layer as the embedding of the image. 
The encoded image is represented as a fixed-length embedding vector $I\in R^{K_I\times d_I}$. To align the image representation with the LLM embedding space, we use an MLP to map the obtained features to the dimensions of LLMs. The mapped embeddings are then concatenated with text embeddings and used as input to LLMs, similar mapping methods are adopted for other modalities. 

\subsubsection{Video Branch}
Considering the inherent information redundancy in videos and memory limitations, we uniformly sample $M$ frames form the video. Each frame is processed by the image encoder, resulting in $V_f=[v_1,v_2,\dots,v_M]$ where $v_i\in R^{K_f \times d_f}$ represents the embedding of the $i$-th frame. To preserve temporal information, we introduce temporal position encoding to the representation. The enhanced representation is then fed into the Video Q-former with the same structure as the Q-Former in BLIP-2~\cite{li2023blip} to aggregate video information, which generates $k_V$ video embedding vectors of dimensions $d_V$. These vectors form the representation $V\in R^{k_V\times d_V}$ for the entire video. 
% The video branch effectively captures both content and temporal information. This enables the model to comprehend visual content while preserving temporal cues for multi-modal grounding tasks.

\subsubsection{Audio Branch}
The audio branch follows a structure similar to the video branch. We employ the  ImageBind audio encoder, which processes 2-second audio clips with a 16kHz sampling rate and converts them into spectrograms using 128 mel-spectrogram bins. We sample $N$ 2-second segments from the original audio and transform each segment into a vector, resulting in $A_s=[a_1,a_2,\dots a_N]$, where $a_i\in R^{K_s\times d_s}$ represents the embedding of the $i$-th aduio segment. 
We incorporate temporal position encoding into $A_s$. Finally, we obtain a fixed-length audio representation sequence denoted as $A\in R^{k_A\times d_A}$ using the audio Q-former like video branch. 

\subsubsection{Spatial-temporal Representation}\label{representation}
We represent the bounding box in an image using four relative coordinate values: $[x_1,y_1,x_2,y_2]$. These values correspond to the upper left corner point and the lower right corner point of the bounding box. Each value is rounded to three decimal places. We concatenate this textual representation after the description related to the bounding box. 
Similarly, for representing timestamps, we use two two-digit decimals $\{ t_1, t_2 \}$ to indicate the relative values of the start and end times of a time segment with respect to the total duration.
This representation allows us to train the model without requiring additional vocabulary expansion or training. Examples of the training dataset are shown in Figure~\ref{fig:dataset}.
% By incorporating both spatial and temporal information in this way, the model develops a comprehensive understanding of fine-grained details. 

\subsection{Coarse-to-Fine Training and Dataset}\label{sec:dataset}
% Different from the previous typical two-stage training methods consisting of feature alignment pre-trainding and instruction tuning, we introduces an additional stage to
% improve the cross-modal understanding ability for the model.
% Specifically, the first stage, feature alignment, aims to train
% the visual adapter, to align video features with LLM’s semantic space. The second stage, boundary perception, focuses on enabling LLM to develop attentional capabilities
% for specific moments, facilitating the understanding of various events occurring within the video. The third stage, instruction tuning, allows LLM to align with human intent
% and enabling more precise event localization and description. In the following sections, we will elaborate on the
% training methods and datasets utilized for each of these three
% stages.
We employ a three-stage coarse-to-fine training strategy to train the model, while constructing specific datasets for each stage.
% In this section, we introduce our three-stage coarse-to-fine training strategy. We will present the stage-specific dataset construction pipeline.
% Through this training approach, our model achieves a higher level of semantic understanding and grounding ability.

\subsubsection{Multi-modal Pre-training}
This stage focus on enabling the model to comprehend multi-modal inputs and develop a high-level semantic perception of the input. 
During the training process, the LLM and the encoders for each modality remain frozen, while only the adapters for each modality are trained. 
% Training is conducted with a batch size of 64, a learning rate of 2e-3, and is completed in approximately 10 hours using 8 A100 GPUs for GroundingGPT-7B.
\paragraph{Training Dataset}
We utilize public pretraining datasets as the primary source of our data. The training data for the image and video modalities is LLaVA-Pretrain-595k and Valley-Pretrain-703k, respectively. To construct the audio data, we adopt a similar approach as in LLaVA, leveraging the Wavcaps~\cite{mei2023wavcaps} dataset. Each sample is accompanied by a sampled instruction that requires the model to provide a concise description of the audio to construct a single-turn conversation.

\subsubsection{Fine-grained Alignment Tuning}
 The second stage aims to enable the model to comprehend more detailed information, including coordinates and timestamps. Through training in this stage, the model achieves impressive results in various grounding tasks, establishing a more comprehensive and refined understanding ability. During the training process, the encoders for each modality are frozen, while the LLM and adapters are trained. 
% Training is performed with a batch size of 32, a learning rate of 2e-5, and takes around 40 hours using 8 A100 GPUs for GroundingGPT-7B.

\paragraph{Training Dataset}
 The training data used in this stage includes the spatial-temporal representation mentioned in Section \ref{representation}. To address the scarcity of fine-grained multi-modal data, we construct a multi-modal dataset specifically designed for this stage. The dataset is primarily obtained by converting publicly available datasets. 
 As depicted in the  left part of Figure~\ref{fig:dataset}, task descriptions are provided to GPT-3.5 to generate a task-specific question pool. For each data sample, a question is randomly selected from the pool, and templates are used to convert the sample's format, resulting in a single-turn conversation. 
 To ensure the quality of the dataset, we carefully select the data by eliminating samples that do not conform to the desired format or criteria.
 For the image modality, we utilize visual grounding datasets such as RefCOCO~\cite{kazemzadeh2014referitgame}, RefCOCO+~\cite{kazemzadeh2014referitgame}, RefCOCOg~\cite{mao2016generation} and Visual Genome~\cite{krishna2017visual} to construct the datasets.
 % To enhance the accuracy of attribute understanding, we have additional tasks such as color recognition, counting, and OCR recognition.
 For the video modality, video temporal grounding datasets such as DiDeMo~\cite{anne2017localizing}, HiREST~\cite{zala2023hierarchical} are utilized for fine-grained alignment. 
 Regarding the sound localization task, we employ the VGGSS~\cite{chen2021localizing} dataset for training. 
 All these datasets are transformed into single-turn conversation format following the aforementioned pipeline for training.

\subsubsection{Multi-granularity Instruction Tuning}\label{sec:stage3data}
After the training in the first two stages, the model has acquired a strong understanding and grounding capability. This stage aims to enable the model to generate responses that better align with human preferences and improve multi-modal interactions. We train the model using instruction-tuning datasets at different granularities.
Similar to the second stage, the encoders for each modality are frozen, while the LLM and adapters are trained.
% The model is trained for one epoch with a batch size of 32 and a learning rate of 1e-5. Training on 8 A100 GPUs for GroundingGPT-7B is completed in approximately 8 hours.

\paragraph{Training Dataset}
 The data utilized in this stage consists of high-quality fine-grained instruction-tuning dataset we construct and public instruction-tuning dataset. 
 As illustrated in the right part of Figure~\ref{fig:dataset}, we select a subset of public datasets for human annotation to create in-context examples. It assists in guiding GPT-3.5 to follow similar patterns when generating instruction-tuning dataset. Subsequently, task-specific system prompts and randomly selected examples are input to GPT-3.5 to generate single/multi-turn conversations. For the image modality, we construct fine-grained datasets using the Flickr30K Entities~\cite{plummer2015flickr30k} dataset, including detailed descriptions and conversations. To enhance the model's fine-grained reasoning capability, we utilize the VCR~\cite{zellers2019recognition} dataset to construct a reasoning dataset with coordinates. For the video modality, we constructed datasets with temporal information by incorporating datasets from various video tasks such as DiDeMo~\cite{anne2017localizing} and Activitynet Captions~\cite{krishna2017dense}, along with other relevant sources. 
 The public instruction-tuning datasets we use include LLaVA-v1.5-mix665k, Valley-Instruct-73k, Videochat-Instruct-11k, and an audio instruction-tuning dataset constructed using Clotho~\cite{drossos2020clotho} dataset.
 For more details about the datasets, please refer to appendix~\ref{sec:appendix_data}.

 During training, in order to prevent catastrophic forgetting in subsequent training stages, we adopt a sampling strategy that incorporates training data from previous stages. The training process employs a consistent training objective as follows:
\begin{equation*}
\begin{aligned}
    L(\theta) = & -\mathbb{E}_{(x, y) \sim D_{\text{current}}} [\log p(y | x)] \\
    & - \alpha \cdot \mathbb{E}_{(x, y) \sim D_{\text{previous}}} [\log p(y | x)],
\end{aligned}
\end{equation*}
where $D_{current}$ denotes the dataset in current training stage, $D_{previous}$ denotes the dataset in previous training stage and $\alpha$ denotes the sampling rate. In the first training stage, $\alpha$ is set to 0.

\section{Experiments}

\subsection{Experimental Setup}

\begin{table*}[t]
    \centering
    \begin{tabular}{cc|ccc|ccc|cc}
    \toprule
    \multirow{2}{*}{Models} & \multirow{2}{*}{LLM Size} & \multicolumn{3}{c}{RefCOCO} & \multicolumn{3}{c}{RefCOCO+} & \multicolumn{2}{c}{RefCOCOg}\\
    \cline{3-10}  & & val & testA & testB & val & testA & testB & val & test \\
    \midrule
    UNITER     & -  & 81.41 & 87.04 & 74.17 & 75.90 & 81.45 & 66.70 & 74.02 & 68.67\\
    MDETR      & -  & 86.75 & 89.58 & 81.41 & 79.52 & 84.09 & 70.62 & 81.64 & 80.89\\
    UniTAB     & -  & 86.32 & 88.84 & 80.61 & 78.70 & 83.22 & 69.48 & 79.96 & 79.97\\
    \midrule
    KOSMOS-2   & 1.6B  & 52.32 & 57.42 & 47.26 & 45.48 & 50.73 & 42.24 & 60.57 & 61.65\\ 
    Shikra     & 7B  & 87.01 & 90.61 & 80.24 & 81.60 & 87.36 & 72.12 & 82.27 & 82.19\\
    NExT-Chat* & 7B & 85.50  & 90.00  & 77.90  & 77.20  & 84.50  & 68.00  & 80.10  & 79.80 \\
    Ferret*    & 7B & 87.49 & 91.35 & 82.45 & 80.78 & \textbf{87.38} & 73.14 & \textbf{83.93} & \textbf{84.76}\\
    \midrule
    % GroundingGPT(224)    & 83.85 & 89.07 & 81.90 & 79.55 & 86.63 & 71.73 & 80.96 & 81.52\\ 
    % GroundingGPT(336)    & 86.88 & 90.85 & 82.22 & 80.74 & 86.79 & 72.08 & 81.29 & 83.05  \\
    GroundingGPT & 7B & \textbf{88.02} & \textbf{91.55} & \textbf{82.47} & \textbf{81.61} & 87.18 & \textbf{73.18} & 81.67 & 81.99  \\
    % GroundingGPT-13B     & 88.54 & 92.20 & 82.75 & 82.61 & 88.16 & 74.19 & 82.56 & 83.50 \\        
    \bottomrule
    \end{tabular}
    \caption{Performance comparison on the referring expression comprehension(REC) task. "*" indicates that the model employs additional image region perception modules.}
    \label{tab:refcoco}
\end{table*}
\begin{table}[t]
    \centering
    \setlength{\tabcolsep}{1mm}{
    \begin{tabular}{c|cc}
    \toprule
    \multirow{2}{*}{Models} & \multicolumn{2}{c}{Charades-STA} \\
    \cline{2-3}
                & R@1(IoU=0.5) & R@1(IoU=0.7)\\
    \midrule
    Video-LLaMA   & 3.8  &  0.9   \\
    VideoChat    & 3.3  &  1.3   \\
    VideoChatGPT & 7.7  & 1.7   \\
    \midrule
    GroundingGPT         & \textbf{29.6} & \textbf{11.9}  \\
    \bottomrule
    \end{tabular}}
    \caption{Performance comparison on the temporal grounding task. All the models have the same LLM size of 7B. }
    \label{tab:vtg}
\end{table}
We employ Vicuna-v1.5~\cite{chiang2023vicuna} as the language model. 
% In the first training stage, the learning rate was set to 1e-3, while in the subsequent two stages, it was set to 2e-5. 
Each training stage lasts for one epoch. During the training process, all images were padded to a square shape and resized to a resolution of $336\times 336$. For each video, 64 frames were sampled, and for each audio, three 2-second segments were sampled and processed. For more details on the hyper-parameter settings, please refer to the appendix~\ref{sec:appendix_hyper}.

\subsection{Quantitative Evaluation}
We conducted extensive experiments for the effectiveness of GroudingGPT in multi-modal grounding and understanding tasks.
% local understanding (multi-modal grounding) and general understanding (multi-modal comprehension) . 

\subsubsection{Multi-modal Grounding}
In this section, we demonstrate that our model achieves impressive fine-grained understanding of multi-modal inputs on grounding tasks. 

\paragraph{Image Grounding}
To assess the image grounding capability of the GroundingGPT model, we conduct experiments on the widely used Reference Expression Understanding (REC) task. The REC task requires the model to locate the bounding box corresponding to a given text reference expression. Our experiments involve three datasets: RefCOCO, RefCOCO+ and RefCOCOg. 
The baselines used for comparing include previous end-to-end multi-modal models UNITER~\cite{chen2020uniter}, MDETR~\cite{kamath2021mdetr}, UniTAB~\cite{yang2022unitab}, and the LLM-based multi-modal grounding models KOSMOS-2, Shikra, NExT-Chat and Ferret.
For GroundingGPT model, we use a unified prompt like "Output the coordinate of <exp>", where "<exp>" represents the reference expression. 
% A predicted bounding box is considered correct if the intersection-over-union (IoU) between the predicted bounding box and the GT box is greater than 0.5. 
The results on the REC task is presented in Table \ref{tab:refcoco}. GroundingGPT demonstrates remarkable performance across multiple datasets and performs comparably to specialized fine-tuned models or MLLMs that incorporate additional image region perception modules.

\paragraph{Video Grounding}
To evaluate the video grounding capability of GroundingGPT, we conduct experiments on the temporal video grounding task. For the task, we employed datasets from Charades-STA~\cite{gao2017tall}. 
The predicted time segments are compared with the corresponding ground truth time segments to calculate the IoU.
The evaluation metric used is "R@1, IoU = m", which measures the percentage of correctly retrieved moments with an IoU greater than m. We set the values of m as {0.5, 0.7} to assess different levels of accuracy.
As shown in Table \ref{tab:vtg}, GroundingGPT exhibits excellent performance in temporal video grounding task compared to previous video MLLMs, which primarily focuses on entire video understanding.

\subsubsection{Multi-modal Understanding}
\begin{table*}[t]
    \centering

    \setlength{\tabcolsep}{1mm}  {
    \begin{tabular}{cc|ccccc|cccc}
    \toprule
    Models & LLM Size & $\text{VQA}^{\text{v2}}$ & GQA & VisWiz & $\text{SQA}^{\text{I}}$ & $\text{VQA}^{\text{T}}$ & POPE & MME & MMB & $\text{LLaVA}^{\text{W}}$\\
    \midrule
    BLIP-2          & 13B    & 41.0  & 41    & 19.6  & 61    & 42.5  & 85.3  & 1293.8    & -     &38.1\\
    InstructBLIP    & 7B     & -     & 49.2  & 34.5  & 60.5  & 50.1  & -     & -         & 36    &60.9\\
    InstructBLIP    & 13B    & -     & 49.5  & 33.4  & 63.1  & 50.7  & 78.9  & 1212.8    & -     &58.2\\ 
    Shikra          & 13B    & 77.4  & -     & -     & -     & -     & -     & -         & 58.8  & - \\ 
    % Qwen-VL         & Qwen-7B       & 78.8  & 59.3  & 35.2  & 67.1  & 63.8  & -     & -         & 38.2  & - \\ 
    % Qwen-VL-Chat    & Qwen-7B       & 78.2  & 57.5  & 38.9  & 68.2  & 61.5  & -     & 1487.5    & 60.6  & - \\ 
    LLaVA-1.5       & 7B     & 78.5  & 62.0  & 50.0  & 66.8  & \textbf{58.2}  & 85.9  & \textbf{1510.7}   & \textbf{64.3}  &63.4\\ 
    \midrule
    GroundingGPT    & 7B     & \textbf{78.7}  & \textbf{62.1}  & \textbf{55.1}  & \textbf{78.7}  & 55.2  & \textbf{87.4}  & 1454.2    & 63.8  &\textbf{70.9}\\ 
    \bottomrule
    \end{tabular}}
    \caption{Comparison of MLLMs on image understanding benchmarks. Benchmark names are abbreviated due to space limits. VQA-v2~\cite{goyal2017making}; GQA~\cite{hudson2019gqa}; VisWiz~\cite{gurari2018vizwiz}; $\text{SQA}^{\text{I}}$:ScienceQA-IMG~\cite{lu2022learn}; $\text{VQA}^{\text{T}}$: TextVQA~\cite{singh2019towards}; POPE~\cite{li2023evaluating}; MME~\cite{fu2023mme}; MMB:MMBench~\cite{liu2023mmbench}; $\text{LLaVA}^{\text{W}}$: LLaVA-Bench (In-the-Wild)~\cite{liu2023visual}. }
    \label{tab:image_understanding}
\end{table*}
\begin{table*}[!t]
    \centering
    \setlength{\tabcolsep}{1mm} {
    \begin{tabular}{cc|cc|cc|cc}
    \toprule
    \multirow{2}{*}{Models} & \multirow{2}{*}{LLM Size} & \multicolumn{2}{c}{MSVD-QA} & \multicolumn{2}{c}{MSRVTT-QA} & \multicolumn{2}{c}{ActivityNet-QA}\\
    \cline{3-8}
    & & Accuracy & Score & Accuracy & Score & Accuracy & Score \\
    \midrule
    % FrozenBiLM      & 32.2  & -     & 16.8  & -     & 24.7  & -\\
    VideoChat     &7B  & 56.3  & 2.8   & 45.0  & 2.5   & 26.5   & 2.2\\
    Video-LLaMA    &7B & 51.6  & 2.5   & 29.6  & 1.8   & 12.4  & 1.1\\
    Video-ChatGPT  &7B & 64.9  & 3.3   & 49.3  & 2.8   & 35.2  & 2.7\\
    Valley  &7B & 65.4  & 3.4   & 45.7  & 2.5   & 42.9  & 3.0\\
    % Chat-UniVi      & 65.0  & 3.6   & 54.6  & 3.1   & 45.8  & 3.2\\
    \midrule
    GroundingGPT  &7B  & \textbf{67.8}  & \textbf{3.7}   & \textbf{51.6}  & \textbf{3.1}   & \textbf{44.7}  & \textbf{3.2}\\
    \bottomrule
        \end{tabular}}
    \caption{ Comparison of MLLMs on video understanding benchmarks. We adopt the evaluation methodology in Video-ChatGPT~\cite{maaz2023video} for evaluation.}
    \label{tab:video_understanding}
\end{table*}
\begin{table*}[!]
    \centering
    \setlength{\tabcolsep}{0.4mm}{
    \begin{tabular}{cc|ccc|ccc|ccc}
    \toprule
    \multirow{2}{*}{Models} & \multirow{2}{*}{LLM Size} & \multicolumn{3}{c}{Random} & \multicolumn{3}{c}{Popular} & \multicolumn{3}{c}{Adversarial}\\
    \cline{3-11}
    & & Accuracy & F1-Score & Yes & Accuracy & F1-Score & Yes & Accuracy & F1-Score & Yes \\
    \midrule
    LLaVA        & 7B & 72.16  & 78.22 & 76.29 & 61.37 & 71.52 & 85.63 & 58.67 & 70.12 & 88.33\\
    % MM-GPT       & 7B & 50.10 & 66.71 & 99.90 & 50.00 & 66.67 & 100.00 & 50.00 & 66.67 & 100.00\\
    mPLUG-Owl    & 7B & 53.97 & 68.39 & 95.63 & 50.90 & 66.94 & 98.57 & 50.67 & 66.82 & 98.67\\
    MiniGPT-4    & 13B& 79.67 & 80.17 & 52.53 & 69.73 & 73.02 & 62.20 & 65.17 & 70.42 & 67.77\\
    InstructBLIP & 13B& 88.57 & \textbf{89.27} & 56.57 & 82.77 & 84.66 & 62.37 & 72.10 & 77.32 & 73.03\\ 
    Shikra       & 7B & 86.90 & 86.19 & 43.26 & 83.97 & 83.16 & 45.23 & 83.10 & 82.49 & 46.50\\
    % Ferret       & 7B & 90.24 & 89.76 & 43.78 & 84.90 & 84.21 & 45.63 & 82.36 & 82.00 & 48.18\\
    % Chat-UniVi   & 7B & 85.19 & 86.05 & 54.67 & 69.50 & 74.39 & 69.10 & 64.97 & 71.54 & 73.10\\
    \midrule
    GroundingGPT & 7B & \textbf{89.79}  & 89.22 & \textbf{43.13} & \textbf{88.23} & \textbf{87.38} & \textbf{43.23} & \textbf{86.17} & \textbf{85.50} & \textbf{45.43}\\
    \bottomrule
    \end{tabular}}
    \caption{Results on the POPE benchmark for object hallucination evalaution. "Yes" represents the probability of positive answers to the given question.}
    \label{tab:pope}
\end{table*}
We validate that GroundingGPT can maintain or improve the multi-modal understanding ability by introducing grounding tasks. Especially, it can effectively suppress object hallucination. 

\paragraph{Image Understanding}
We evaluate the image understanding capability of GroundingGPT on five question-answering benchmarks and four recent proposed benchmarks specifically designed for vision instruction tuning. These benchmarks provide a comprehensive assessment of the model's capabilities using diverse evaluation metrics. 
The experimental results presented in Table \ref{tab:image_understanding} demonstrate that GroundingGPT achieves state-of-the-art performance on six benchmarks and remains highly competitive on other three benchmarks. Additionally, GroundingGPT exhibits advanced capabilities compared to larger-scale grounding MLLMs, such as Shikra-13B.
% These results highlight the powerful image understanding abilities of GroundingGPT.

% On the VisWiz benchmark, GroundingGPT improves from 50.0 to 55.1 compared to LLaVA-1.5. VisWiz requires models to output "unable to answer" when they cannot answer a question based on the image content. On this metric, GroundingGPT significantly improves from 67.8 to 84.0 compared to LLaVA-1.5. Additionally, on the Pope benchmark used to evaluate the MLLM illusion, GroundingGPT also improves from 85.9 to 87.4.
\paragraph{Video Understanding}
In Table \ref{tab:video_understanding}, we provide a quantitative assessment of the video question answering capabilities of MLLMs on
three datasets: MSVD-QA~\cite{chen2011collecting}, MSRVTT-QA~\cite{xu2016msr} and ActivityNet-QA~\cite{yu2019activitynet}. 
% The evaluation method for this task follows that of Video-ChatGPT.
% We present accuracy and score metrics, evaluated using GPT-Assistant. 
GroundingGPT consistently outperforms other baselines, demonstrating its advanced video understanding capabilities. Notably, GroundingGPT surpasses the powerful baseline
of Valley by 2.4\%, 6.9\% and 1.8\% on MSVD-QA, MSRVTT-QA and ActivityNet-QA, respectively. 
% These results provide evidence of GroundingGPT's exceptional ability to comprehend video content and answer questions based on instructions.

\paragraph{Object Hallucination}
% To further illustrate that the introduction of local information can enhance the global comprehension
We conduct an evaluation of MLLMs regarding their object hallucination in Table \ref{tab:pope}.
% The POPE benchmark consists of three sampling strategies: Adversarial, Popular, and Random.  
% Evaluation metrics include accuracy, F1-Score, and the "Yes" metric, which indicates the probability of the model outputting positive answers (lower values are better).
GroundingGPT achieves the highest performance across all three sampling subsets. Notably, GroundingGPT-7B outperforms larger models such as InstructBLIP-13B, on the challenging Adversarial subset, GroundingGPT exhibits 14.07\% increase in accuracy and 8.18\% increase in f1-score, while experiencing a 27.60\% decrease in the "Yes" metric.
% Furthermore, VisWiz benckmark requires models to output "unable to answer" when they cannot answer a question based on the image content. 
Additionally, on the "unanswerable" subset of VisWiz benchmark, GroundingGPT significantly improves from 67.8\% to 84.0\% compared to LLaVA-1.5. This improvement reveals that our model not only acquires a high-level semantic understanding of the overall image but also develops the ability to comprehend local details. This comprehensive understanding of the image enables the model to effectively suppress the occurrence of hallucinations.

\begin{figure*}[t]
    \centering
    \includegraphics[width=0.95\textwidth]{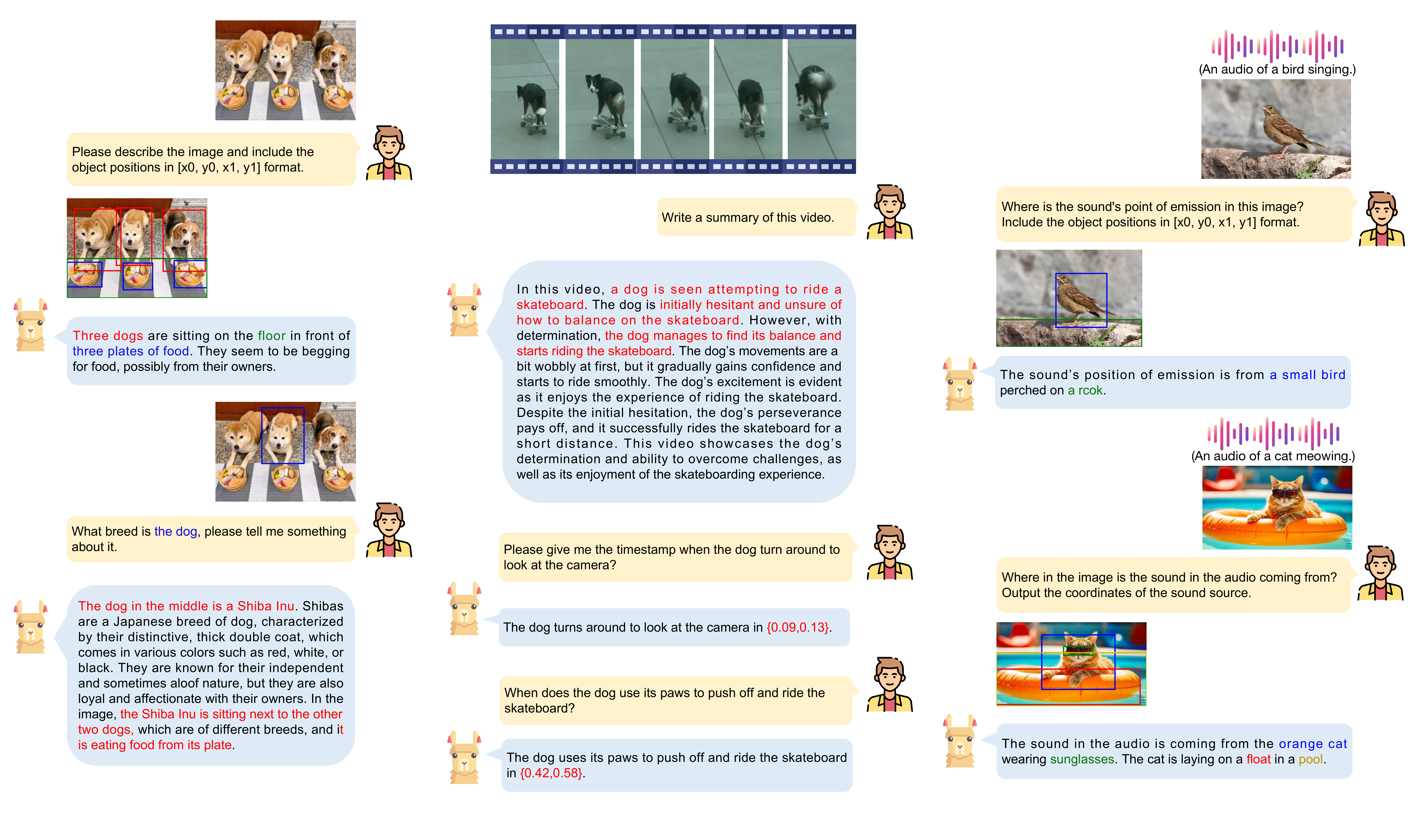}
    \caption{Qualitative results of GroundingGPT on multi-modal grounding tasks.}
    \label{fig:demo}
\end{figure*}

\subsubsection{Qualitative Results}
We provide qualitative results to demonstrate the multi-modal understanding and grounding capabilities of our model. Figure~\ref{fig:demo} showcases examples illustrating the performance of GroundingGPT in multi-modal grounding tasks. 
More examples are available in appendix~\ref{sec:appendix_examples}. 
These results indicate that GroundingGPT excels in fine-grained multi-modal grounding tasks while maintaining a comprehensive understanding of multi-modal inputs. 
% This demonstrates the effectiveness and generality of our model.

\subsection{Ablation Study}
\begin{table}[t]
    \centering
    \begin{tabular}{ccc|ccc}
    \toprule
  \multirow{2}{*}{S1} & \multirow{2}{*}{S2} & \multirow{2}{*}{S3} & \multicolumn{3}{c}{RefCOCO} \\
    \cline{4-6}
     & & & val & testA & testB \\
    \midrule
    C+F & F & C+F & 82.43 & 86.87 & 75.37 \\
      C & F & C+F & \textbf{84.68} & \textbf{88.88} & \textbf{78.94} \\

    \bottomrule
    \end{tabular}
    \caption{Ablation studies of the training strategy on the REC task. The S1 to S3, denoted as stage 1 to stage 3, represent the training data used in each stage. "C" represents coarse-grained data, while "F" represents fine-grained data. To quickly illustrate the performance, we adopt a simple training setting ($224 \times 224$  simage resolution and linear adapter) as the toy model. }
    \label{tab:ablation_step}
\end{table}

% To validate the effectiveness of our approach， We investigate the impact of our training strategy, model architecture, and size on the results using the REC task.
% To better boost the model to capture global semantic information and local fine-grained
% details simultaneously, we  specifically investigate the impact of training strategy, model architecture and size. 
To validate the effectiveness of our approach, we conducted experiments on the REC task to assess the impact of training strategy, model architecture, and size on the results.

\paragraph{Training Strategy}
As shown in Table~\ref{tab:ablation_step}, it is evident that including fine-grained training data in the first stage results in a decline in performance. This can be attributed to the model's limited understanding of the images at this early stage. The introduction of fine-grained data during training may introduce interference and hinder the model's learning. This finding further validates the effectiveness of our coarse-to-fine training strategy.

\begin{table}[t]
    \centering
    \begin{tabular}{cc|ccc}
    \toprule
    \multirow{2}{*}{LLM Size} & \multirow{2}{*}{Adapter} &  \multicolumn{3}{c}{RefCOCO} \\
    \cline{3-5}
       & & val & testA & testB \\
    \midrule
     7B  & Linear  & 86.01 & 90.45 & 80.43 \\  
     7B  & MLP     & 88.02 & 91.55 & 82.47 \\
     13B  & MLP    & \textbf{88.26} & \textbf{92.05} & \textbf{82.65} \\
           
    \bottomrule
    \end{tabular}
    \caption{Ablation studies of the model architecture, LLM size on the REC task.}
    \label{tab:ablation_size}
\end{table}

\paragraph{Model Architecture and Size}
As shown in Table~\ref{tab:ablation_size}, the top two rows demonstrates that replacing the linear layer with an MLP in the adapter leads to performance enhancement. This improvement can be attributed to the preservation of more comprehensive image information and the improved mapping of image embeddings to the LLM space.
Besides, increasing the LLM size leads to an improvement. This can be attributed to the fact that larger language model possess richer knowledge and stronger modeling capabilities.

\section{Conclusion}
In this paper, we introduce GroundingGPT, a unified end-to-end multi-modal grounding model. To the best of our knowledge, this is the first multi-modal large language model capable of performing multi-modal grounding and understanding tasks.
We adopt a three-stage coarse-to-fine training strategy, accompanied by the construction of stage-specific training datasets, to effectively train the model.
Our model demonstrates remarkable performance in multi-modal grounding and understanding tasks. Extensive experiments conducted on a wide range of MLLM benchmarks confirm the effectiveness and generality of our model. To foster further advancements in this field, we make our model, code, and dataset openly accessible.

% Through training on a diverse multi-modal and multi-granularity dataset, GroundingGPT achieves better perception of multi-modal inputs and demonstrates improved performance on tasks requiring fine-grained understanding. To address the scarcity of relevant data, we create a multi-modal grounding dataset encompassing various modalities, tasks, and granularities. To encourage further advancements in this field, we will make our model, code, and dataset openly accessible. In future work, we aim to extend GroundingGPT to accommodate additional input and output modalities while exploring more sophisticated grounding methods.

\section{Limitations}
% \paragraph{Language Hallucination.}
% Similar to previous studies, our model is built upon the pretrained large language model, which may have certain limitations and occasionally exhibit hallucination phenomena. It is possible for the model to generate content that does not exist in the input or provide incorrect knowledge.
\paragraph{Sampling Strategy}
Due to computational memory constraints, GroundingGPT adopts a sampling approach when processing videos and audios. However, this method inevitably results in some loss of crucial information, especially when dealing with longer videos. One future research direction is to explore better modeling approaches for longer videos and minimize information loss.

\paragraph{Cross-modal Inputs}
At present, the majority of the training data primarily consists of single-modal inputs. However, further exploration is needed to address the challenges posed by multi-modal inputs. In the future, we plan to investigate methods for accomplishing grounding tasks in the context of simultaneous multi-modal inputs. For instance, we aim to simultaneously perform spatial and temporal grounding on input videos. Additionally, we will annotate such data to foster advancements in this field.

\paragraph{Grounding Ability}
Despite achieving promising results in multi-modal grounding tasks, GroundingGPT currently lacks the capability to output more fine-grained grounding results such as segmentation masks. In future work, we plan to expand the grounding tasks to support a broader range of grounding requirements.

\bibliography{anthology,custom}
\bibliographystyle{acl_natbib}
\newpage
\appendix
\section{Implementation details}\label{sec:appendix_hyper}
We provide more details of our experiment configuration for
reproducing our model. We provide hyper-parameters for
all stages in Table~\ref{tab:parameter}. 
\begin{table}[h]
    \centering
    \begin{tabular}{c|ccc}
    \toprule
    Settings & Stage1 & Stage 2 & Stage3 \\
    \midrule
    batch size & 64 & 16 & 8 \\
    learning rate & 1e-3 & 2e-5 & 2e-5\\
    learning schedule & \multicolumn{3}{c}{Cosine decay} \\
    warm up ratio & 0.03 & 0.03 & 0.03\\
    weight decay & 0.0 & 0.0 & 0.0 \\
    epoch & 1 & 1 & 1\\
    bf16 & \Checkmark & \Checkmark & \Checkmark\\
    tf32 & \Checkmark & \Checkmark & \Checkmark\\
    grad accumulate & 1 & 2 & 2\\
    DeepSpeed stage & \multicolumn{3}{c}{ZeRO2} \\
    GPUs  & \multicolumn{3}{c}{8×A100} \\
    \bottomrule
    \end{tabular}
    \caption{The hyper-parameters for model training.}
    \label{tab:parameter}
\end{table}
% \section{Dataset Appendix}
% \label{sec:appendix}
\section{Training Dataset Source}\label{sec:appendix_data}
In Table \ref{tab:source}, we provide a comprehensive list of the datasets used in constructing our training dataset. This includes the data utilized in all three stages. It should be noted that a significant portion of the data needs to be constructed in the desired format using publicly available data. Please refer to the section~\ref{sec:dataset} for specific guidance on this matter.

\begin{table*}[h]
    \centering
    \renewcommand\arraystretch{2}
    \setlength{\tabcolsep}{5mm}{
    \begin{tabular}{cc|c}
    \toprule
    Training Stage & Modality & Dataset source \\
    \midrule
    \multirow{3}{*}{Stage1}
        & Image & LLaVA-Pretrain-595k\\
    \cline{2-3}
        & Video & Valley-Pretrain-703k\\
    \cline{2-3}
        & Audio & Wavcaps\\
    \midrule
    \multirow{3}{*}{Stage2} 
        & Image & RefCOCO, RefCOCOg, RefCOCO+, Visual Genome\\
    \cline{2-3}
        & Video & DiDeMo, HiREST, Charades-STA, Didemo\\
    \cline{2-3}
        & Audio & VGGSS\\
    \midrule
    \multirow{3}{*}{Stage3} 
        & Image & LLava-1.5-mix665k, Flickr30k Entities, VCR\\
    \cline{2-3}
        & Video & Valley-Instruct-73k, Videochat-Instruct-11k, Activitynet Captions\\
    \cline{2-3}
        & Audio & Clotho\\
    % \makebox[0.05\textwidth][c]{\textbf{Task}} & \makebox[0.3\textwidth][c]{\textbf{Dataset}}\\
    % \hline
    % Imgae Captioning            & LLaVA-Pretrain-585k \\
    % \hline
    % REC/REG                     & Refcoco, RefCOCOg, Refcoco+, Visual Genome   \\
    % \hline
    % Object Attribute/Relation           & Visual Genome \\
    % % \hline
    % % Object Relation             & Visual Genome \\
    % \hline
    % Image Instruction Tuning    & LLaVA-Instrtuct-150k,VCR \\
    % \hline
    % Video Captioning            & Valley-Pretrain-703k\\
    % \hline
    % Temporal Grounding          & Didemo, Charades-STA, ActivityNet Captions\\
    % \hline
    % Audio Captioning            & Wavecaps \\
    % \hline
    % Video Instruction Tuning    & Valley-Instruct-73k, Videochat-11k \\
    % \hline
    % Audio Instruction Tuning    & Clotho\\
    % \hline
    % Sound Localization          & VGGSS \\
    \bottomrule
    \end{tabular}}
    \caption{The publicly available dataset sources used for constructing the training data.}
    \label{tab:source}
\end{table*}

\section{Dataset Construction Templates}
Table \ref{tab:template} presents the templates utilized for various tasks during the first two training stages. For the sake of demonstration, we provide three examples of instructions for each task.
\begin{table*}[h]
    \centering
    \renewcommand\arraystretch{1.5}
    \begin{tabular}{c|l}
    \hline
    \makebox[0.1\textwidth][c]{\textbf{Task}} & \makebox[0.4\textwidth][c]{\textbf{Template examples}}\\
    \hline
    \multirow{3}{*}{Image Captioning} 
    & Provide a brief description of the given image.  \\ 
    & Write a terse but informative summary of the picture. \\ 
    & Share a concise interpretation of the image provided.\\
    \hline
    \multirow{3}{*}{REG} 
    & What object is present within the specified region<region>?  \\ 
    & Can you identify the item within the region<region>? \\ 
    & Describe the object located within the region<region>.\\
    \hline
    \multirow{3}{*}{REC} 
    & In this image, where is <exp> located?  \\ 
    & Can you identify the position of <exp> within this image? \\ 
    & Please describe the location of <exp> in this image. \\
    \hline
    \multirow{3}{*}{Object Attribute} 
    & What color is this <exp>?  \\ 
    & How many <exp> are visible within this image? \\ 
    & How mang <exp> are there in the image? \\
    \hline
    \multirow{3}{*}{Video Captioning} 
    & Relay a brief, clear account of the video shown.  \\ 
    & Offer a succinct explanation of the footage presented. \\ 
    & Present a compact description of the clip's key features. \\
    \hline
    \multirow{3}{*}{Video Dense Captioning} 
    & Describe the content shown in the video clip<time> of this video.\\ 
    & What can you tell me about the video segment<time> in this video? \\ 
    & Can you provide a description of the video snippet<time>? \\
    \hline
    \multirow{3}{*}{Temporal Grounding} 
    & When did <event> occur in the video? \\ 
    & Tell me the timestamp when <event> happened. \\ 
    & At what time does <event> take place in the video? \\
    \hline
    \multirow{3}{*}{Audio Captioning} 
    & Analyze the audio and provide a description of its content. \\ 
    & Examine the audio and describe the different sounds present. \\ 
    & Provide a detailed summary of the auditory elements in the audio clip. \\
    \hline
    \multirow{3}{*}{Sound Localization} 
    & What is the cause of the sound in this given image? \\ 
    & Can you pinpoint the source of the sound in this image? \\ 
    & Describe the location of the sound's origin in this image. \\
    \hline
    \end{tabular}
    \caption{Instruction templates used to construct the training dataset in the first two stages. The templates include several placeholders: '<region>' represents the coordinates of a region in an image, '<exp>' represents the expression correspond to an image region, '<time>' represents a time segment in a video, and '<event>' represents an event to be located in a video. During the dataset construction process, these placeholders are replaced with corresponding information.}
    \label{tab:template}
\end{table*}

\section{Fine-grained Instruction-tuning Dataset Generation Prompts}
\begin{figure*}[t]
    \centering
    \includegraphics[width=\textwidth]{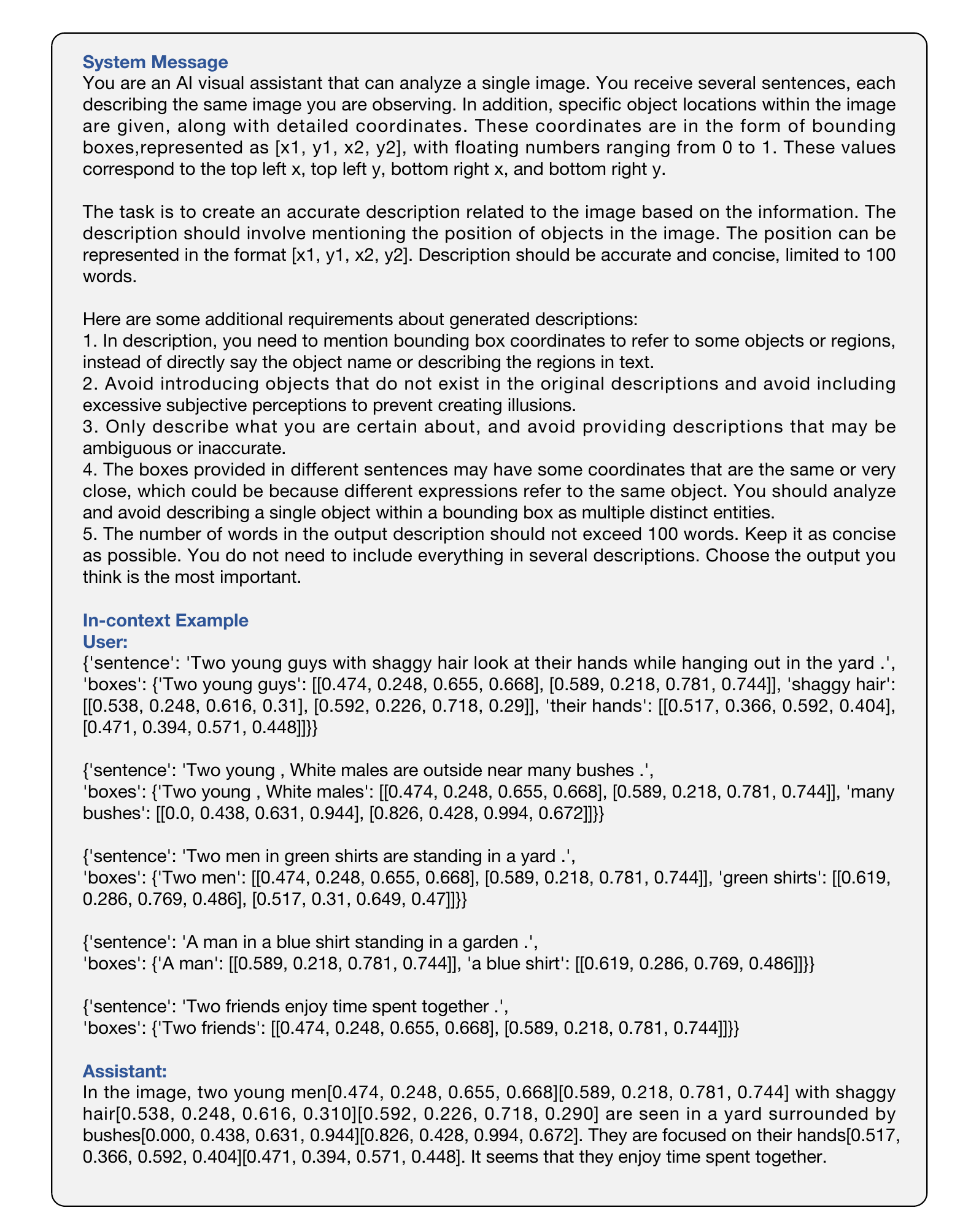}
    \caption{The system message and in-context example used for generating detailed description dataset.}
    \label{fig:prompt1}
\end{figure*}

\begin{figure*}
    \centering
    \includegraphics[width=\textwidth]{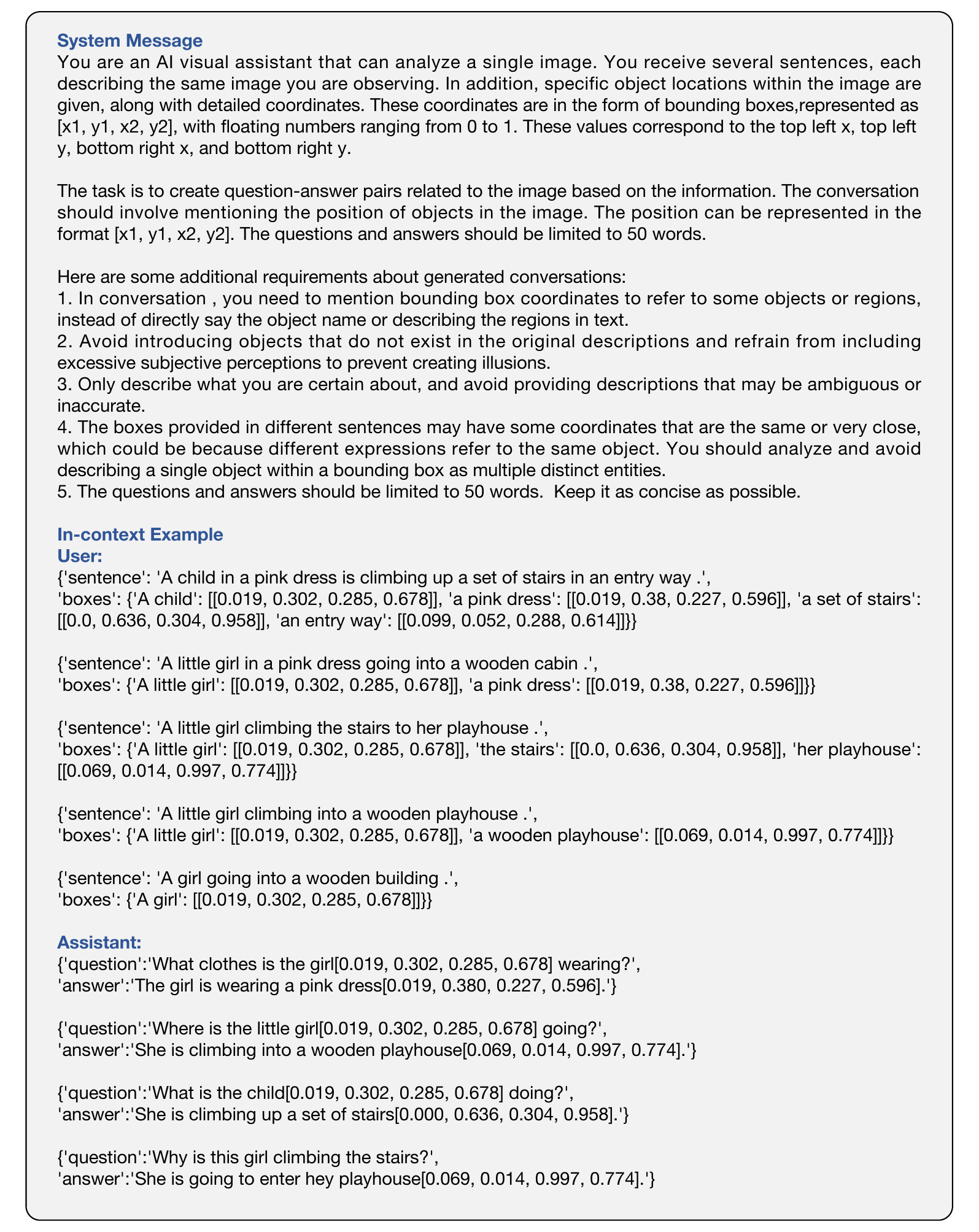}
    \caption{The system message and in-context example used for generating conversation dataset.}
    \label{fig:prompt2}
\end{figure*}

\begin{figure*}
    \centering
    \includegraphics[width=\textwidth]{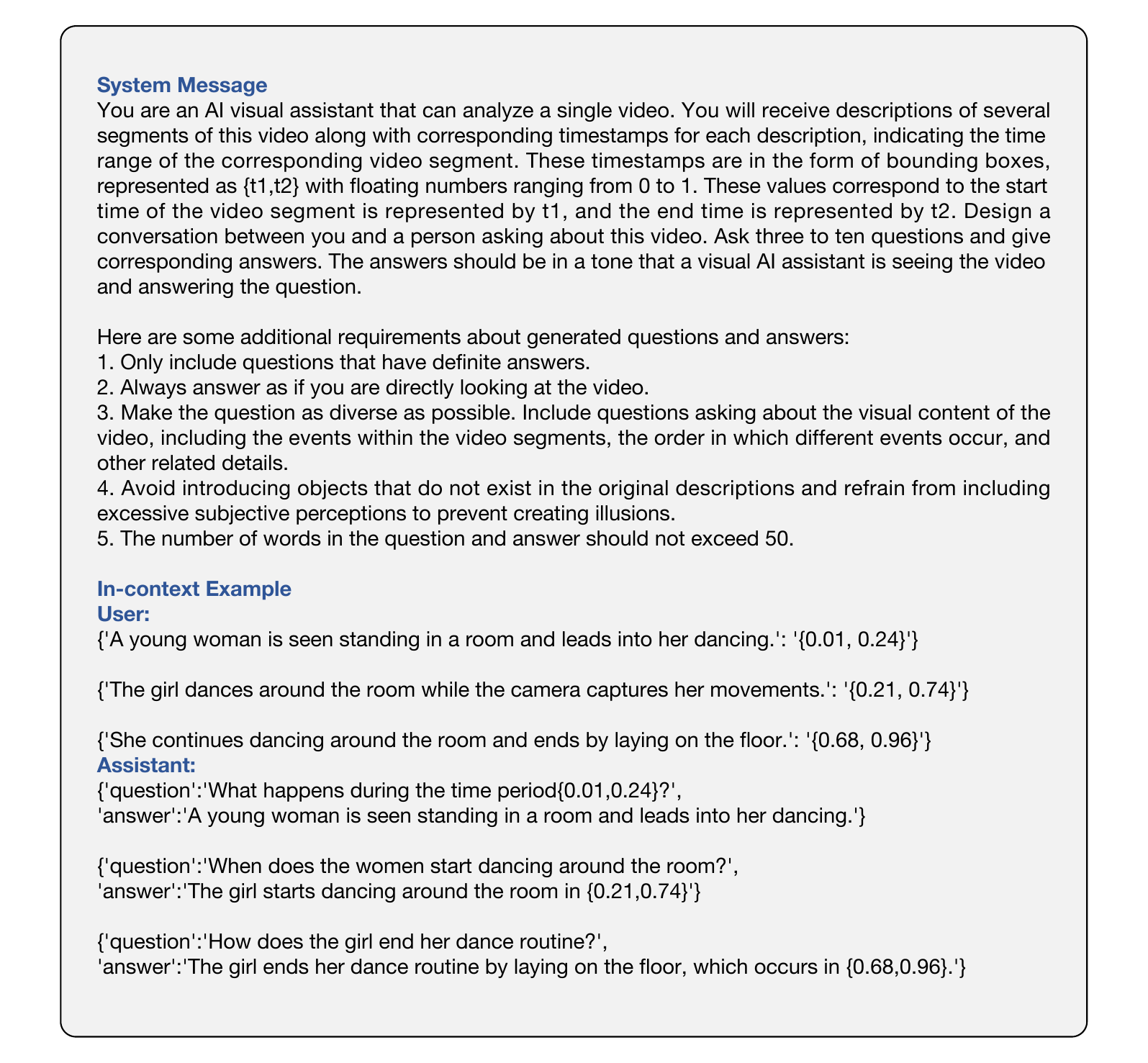}
    \caption{The system message and in-context example used for generating video grounding dataset.}
    \label{fig:prompt3}
\end{figure*}

As shown in section~\ref{sec:stage3data}, we use GPT-3.5 to generate the instruction-tuning dataset. For the image modality, in Figure \ref{fig:prompt1}, we provie the prompt we used to generate the detailed description dataset. In Figure \ref{fig:prompt2}, we provie the prompt we used to generate the conversation dataset. For the video modality, we provie the prompt we used to generate the video grounding instruction-tuning dataset in Figure \ref{fig:prompt3}.

\section{More Visualization}\label{sec:appendix_examples}
\begin{figure*}
    \centering
    \includegraphics[width=0.8\textwidth]{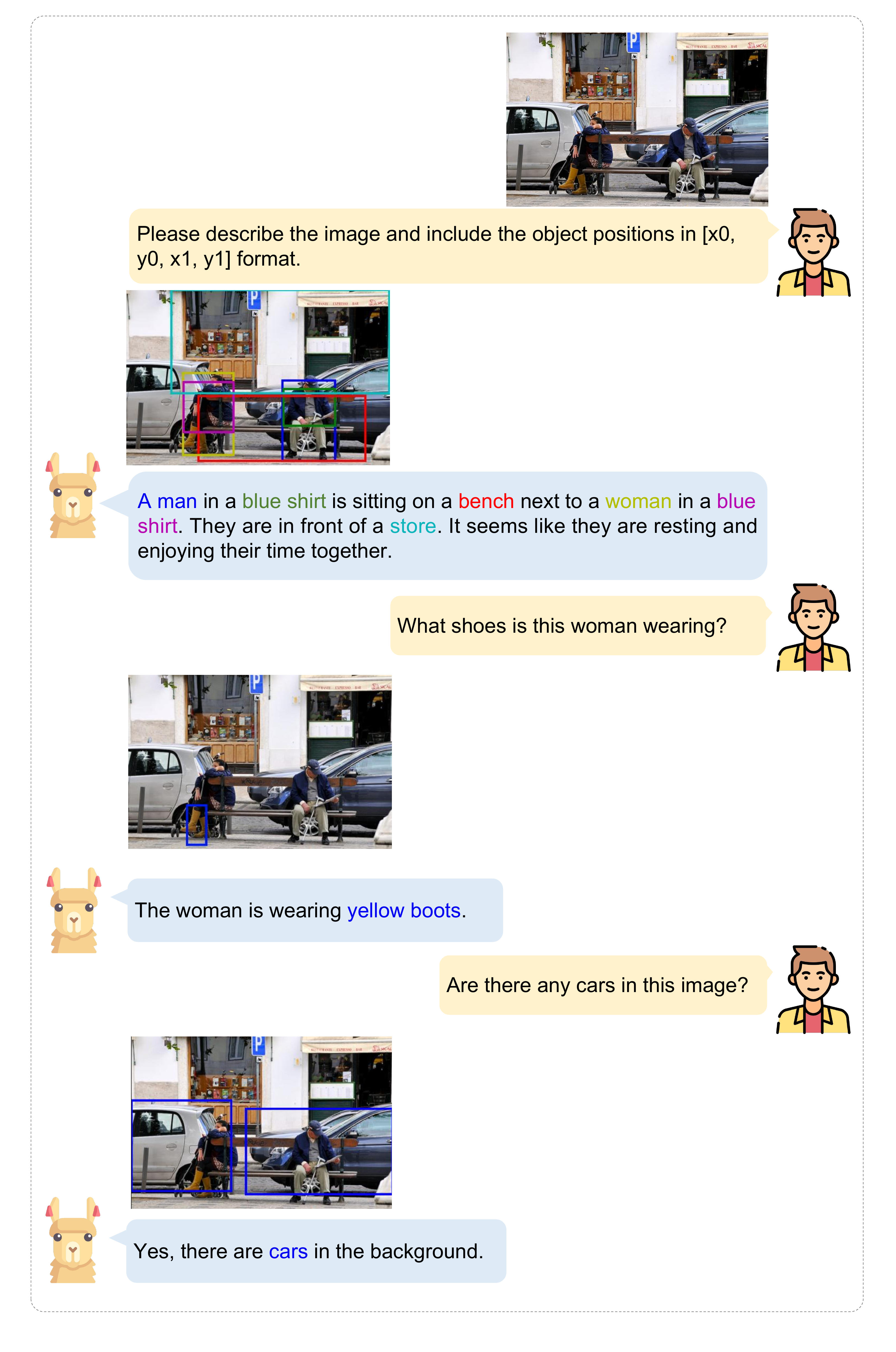}
    \caption{Examples of GroundingGPT's grounding capability in image modality.}
    \label{fig:image_grounding}
\end{figure*}

\begin{figure*}
    \centering
    \includegraphics[width=0.8\textwidth]{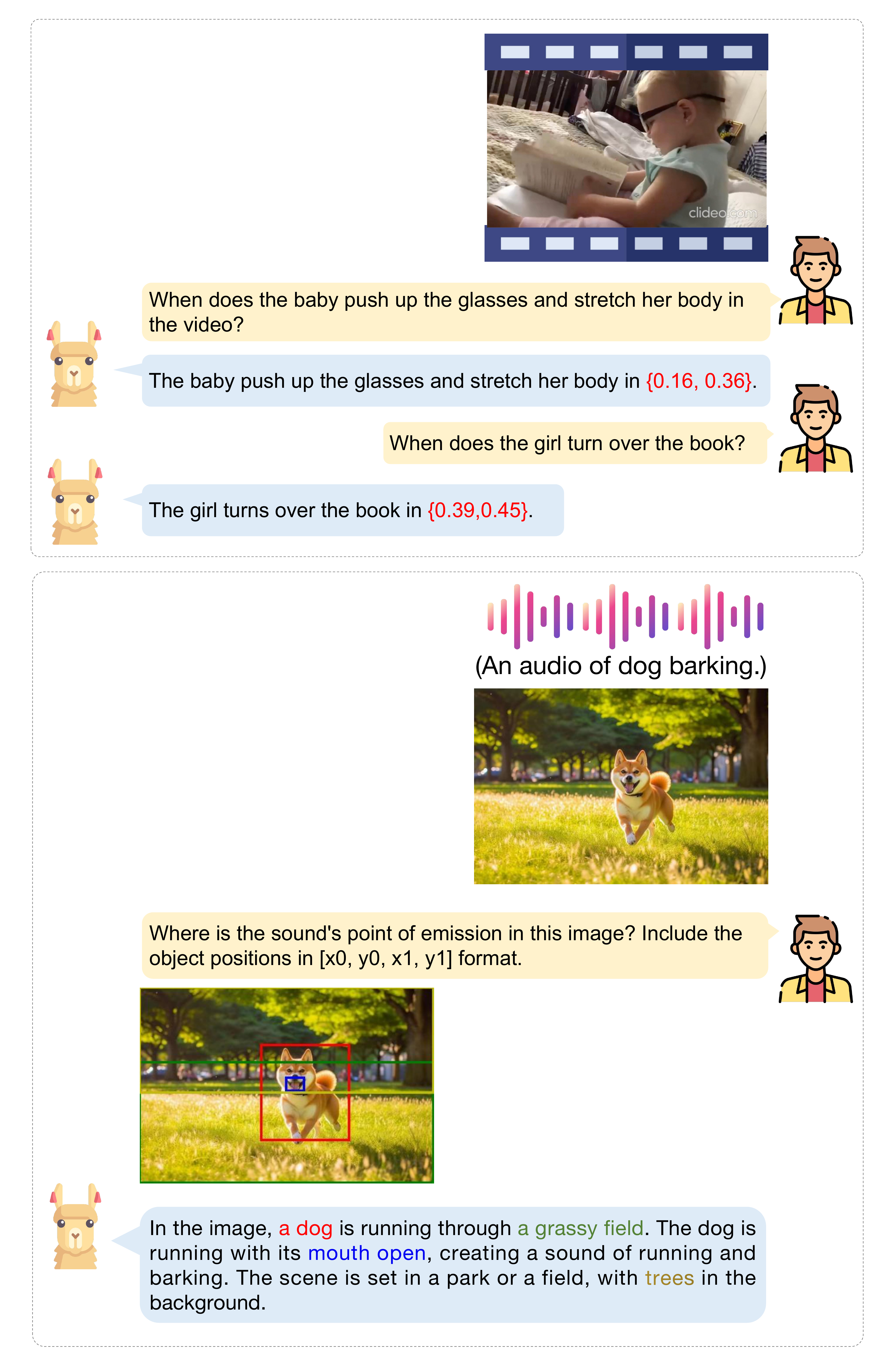}
    \caption{Examples of GroundingGPT's grounding capability in video and audio modality.}
    \label{fig:video_audio_grounding}
\end{figure*}

\begin{figure*}
    \centering
    \includegraphics[width=0.8\textwidth]{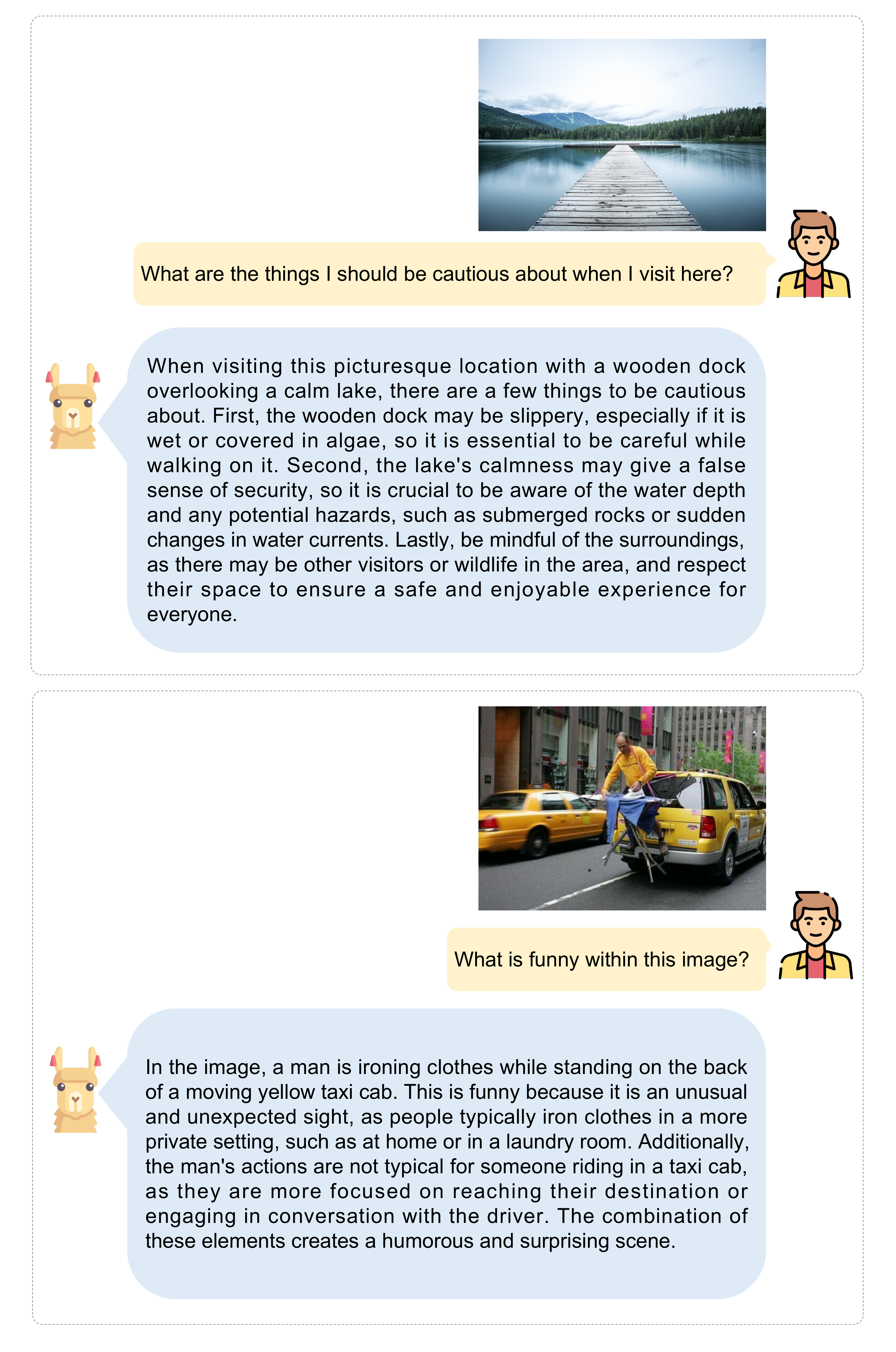}
    \caption{Examples of GroundingGPT's understanding capability in image modality.}
    \label{fig:image_understanding}
\end{figure*}

\begin{figure*}
    \centering
    \includegraphics[width=0.8\textwidth]{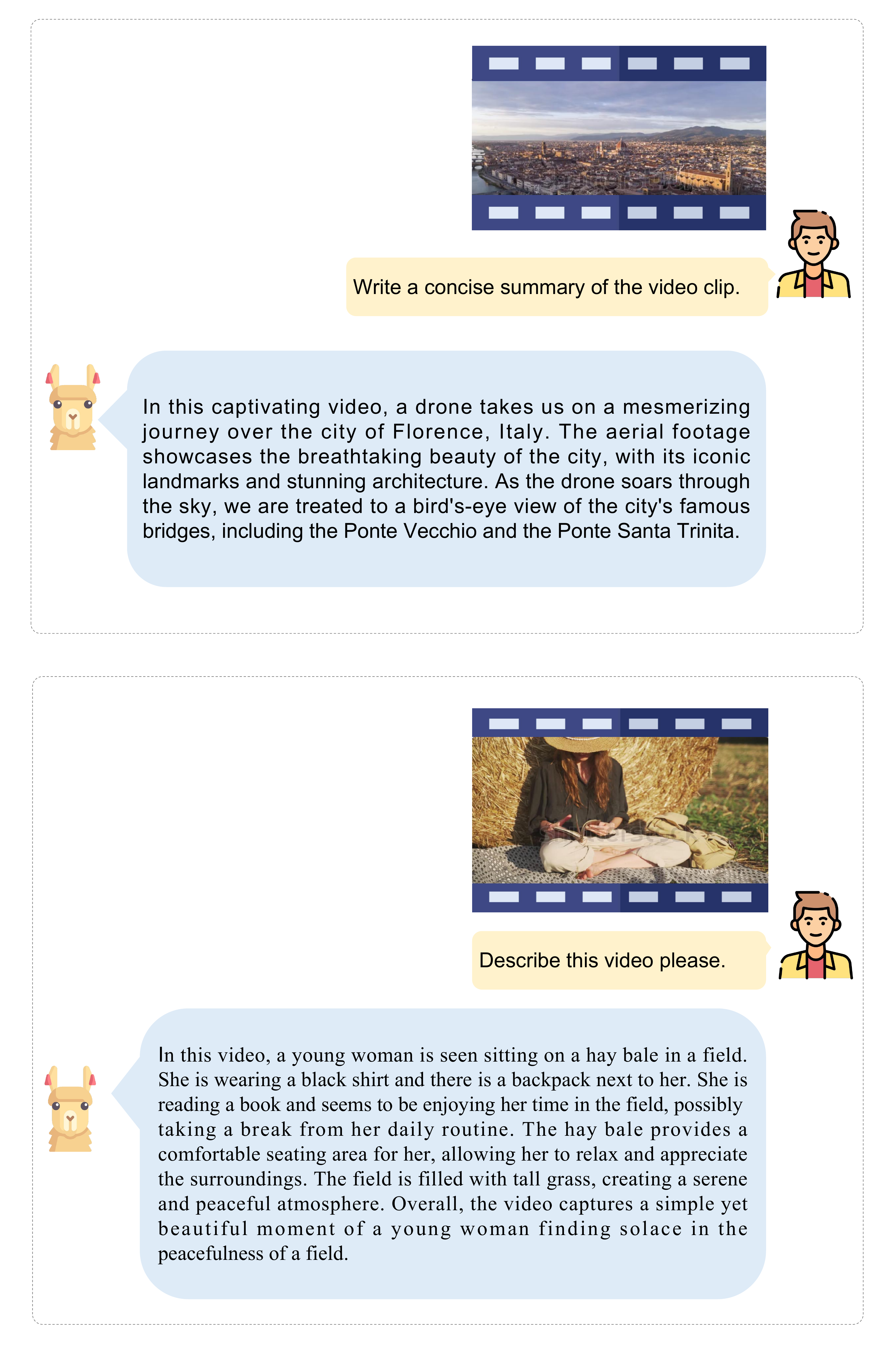}
    \caption{Examples of GroundingGPT's understanding capability in video modality.}
    \label{fig:video_understanding}
\end{figure*}

\begin{figure*}
    \centering
    \includegraphics[width=0.8\textwidth]{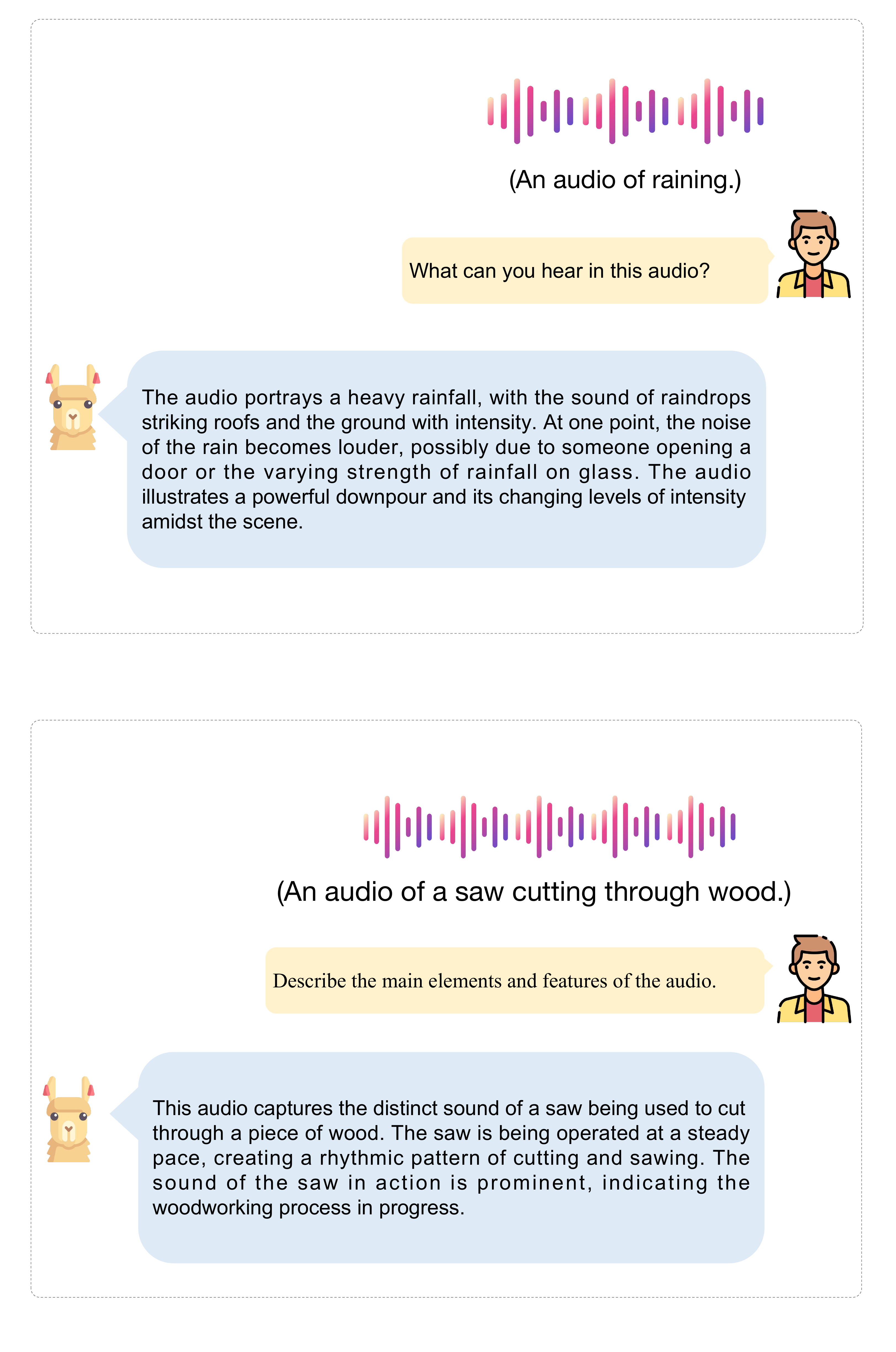}
    \caption{Examples of GroundingGPT's understanding capability in audio modality.}
    \label{fig:audio_understanding}
\end{figure*}

To demonstrate the performance of GroundingGPT in multi-modal grounding and understanding tasks, we present more visualizations in this section. Figure~\ref{fig:image_grounding} and Figure~\ref{fig:video_audio_grounding} showcase the capability of the GroundingGPT model in multi-modal grounding tasks. 
Figure~\ref{fig:image_understanding}, Figure~\ref{fig:video_understanding} and Figure~\ref{fig:audio_understanding} present the capability of GroundingGPT model in multi-modal understanding tasks.

% For more examples, please refer to our project page: \url{https://lzw-lzw.github.io/GroundingGPT.github.io/}, where we provide additional illustrations and demonstrations.

% \begin{figure*}
%     \centering
%     \includegraphics[width=0.8\textwidth]{Figures/demonstration_vg.pdf}
%     \caption{Examples of GroundingGPT's grounding capability in video modality.}
%     \label{fig:video_grounding}
% \end{figure*}

% \begin{figure*}
%     \centering
%     \includegraphics[width=0.8\textwidth]{Figures/demonstration_ag.pdf}
%     \caption{Examples of GroundingGPT's grounding capability in audio modality.}
%     \label{fig:audio_grounding}
% \end{figure*}

\end{document}